\documentclass[journal]{IEEEtran}

\usepackage{xcolor,soul,framed} 

\colorlet{shadecolor}{yellow}
\usepackage[pdftex]{graphicx}
\graphicspath{{../pdf/}{../jpeg/}}
\DeclareGraphicsExtensions{.pdf,.jpeg,.png}

\usepackage[cmex10]{amsmath}
\usepackage{array}
\usepackage{mdwmath}
\usepackage{mdwtab}
\usepackage{tabularx}
\usepackage{eqparbox}
\usepackage{url}
\usepackage{multirow}
\usepackage{booktabs}
\begin{document}
\bstctlcite{}
    \title{VarGes: Improving Variation in Co-Speech 3D Gesture Generation via StyleCLIPS}
  \author{Ming Meng$^{1}$, Ke Mu$^{\ast,1}$, Yonggui Zhu$^{1}$, Zhe Zhu$^{2}$, Haoyu Sun$^{3}$, Heyang Yan$^{1}$ and Zhaoxin Fan$^{\dag,4}$

  \thanks{This Manuscript was accepted by Computational Visual Media on February 13, 2025.}
  \thanks{$ \ast \quad \text{Equal contribution.} $}
  \thanks{\dag \quad Corresponding author.}
  \thanks{1  School of Data Science and Media Intelligence, Communication University of China}
  \thanks{2 Samsung Research America}%
  \thanks{3 Hainan International College, Communication University of China}
  \thanks{4  Beijing Advanced Innovation Center for Future Blockchain and Privacy Computing, Institute of Artificial Intelligence, Beihang University, Beijing, China \and 
Hangzhou International Innovation Institute, Beihang University.}}

\markboth{
}{Roberg \MakeLowercase{\textit{et al.}}: High-Efficiency Diode and Transistor Rectifiers}

\maketitle

\begin{abstract}
Generating expressive and diverse human gestures from audio is crucial in fields like human-computer interaction, virtual reality, and animation. Though existing methods have achieved remarkable performance, they often exhibit limitations due to constrained dataset diversity and the restricted amount of information derived from audio inputs. To address these challenges, we present VarGes, a novel variation-driven framework designed to enhance co-speech gesture generation by integrating visual stylistic cues while maintaining naturalness. Our approach begins with the Variation-Enhanced Feature Extraction (VEFE) module, which seamlessly incorporates {style-reference} video data into a 3D human pose estimation network to extract StyleCLIPS, thereby enriching the input with stylistic information. Subsequently, we employ the Variation-Compensation Style Encoder (VCSE), a transformer-style encoder equipped with an additive attention mechanism pooling layer, to robustly encode diverse StyleCLIPS representations and effectively manage stylistic variations. Finally, the Variation-Driven Gesture Predictor (VDGP) module fuses MFCC audio features with StyleCLIPS encodings via cross-attention, injecting this fused data into a cross-conditional autoregressive model to modulate 3D human gesture generation based on audio input and stylistic clues. The efficacy of our approach is validated on benchmark datasets, where it outperforms existing methods in terms of gesture diversity and naturalness. The code and video results will be made publicly available upon acceptance:
\url{https://github.com/mookerr/VarGES/}.
\end{abstract}


%
\IEEEpeerreviewmaketitle


\section{Introduction}

\IEEEPARstart{H}{ead}, hand, and body gestures are essential components of human communication, playing a pivotal role in augmenting linguistic expression, conveying emotions and attitudes, and facilitating the coordination of dialogue  \cite{1,2,3}. With the increasing use of virtual characters and robots across diverse domains such as education, entertainment, and medicine \cite{4,5}, generating natural and contextually appropriate gestures based on speech has become a significant research challenge. This challenge encompasses multiple disciplines, including computer vision, natural language processing, and human-computer interaction, among others. Furthermore, it finds application in various scenarios, such as virtual hosts, intelligent assistants, and social robots.

The task of generating head, hand, and body gestures in synchronization with speech can be broadly classified into three primary methodologies: rule-based approaches\cite{6,7,18}, statistical model-based techniques\cite{8,9}, and learning-based methods\cite{10,11,12,13,14,57}. Learning-based methods have notably emerged as the forefront of this field, showcasing remarkable proficiency in producing gestures that are both fluid and natural, thereby capturing the intricate dynamics of human expressiveness. These methods have set a high standard by effectively aligning speech with gesture nuances. Nevertheless, achieving a broad spectrum of diverse 3D human gestures remains a significant challenge. Recent strides have been made by incorporating speaker identity to enrich gesture variation \cite{11,13,15,16,17}. Despite these advancements, these methodologies often find themselves constrained by datasets restricted to specific figures, thereby limiting the breadth of gesture styles they can generate. This leads to the learning of fixed patterns, which constrains the variability of gestures across different speech inputs. This inherent limitation fuels our motivation to develop innovative frameworks that push beyond these existing boundaries, striving for greater diversity and adaptability in gesture generation.
\begin{figure*}
\centering
\includegraphics[width=1\linewidth]{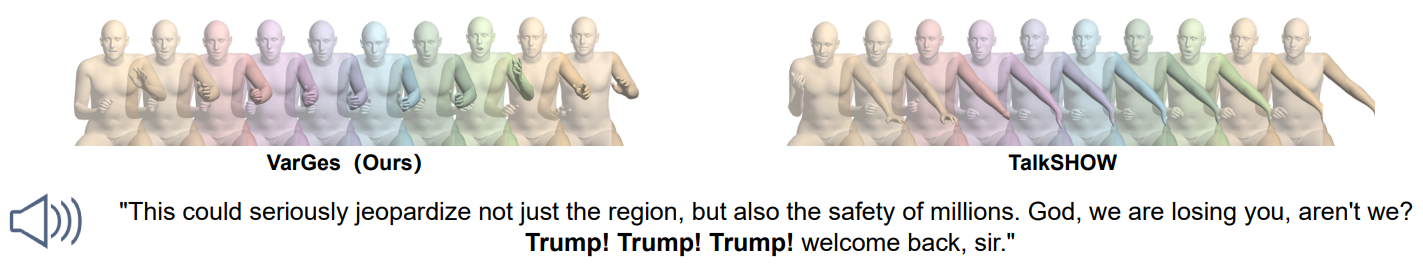}
\caption{Examples of our generated gestures video frames. Compared with TalkShow, our method shows a richer variety in generating character gestures, and the gestures are more natural and smooth.}
\label{fig1} 
\end{figure*}

To address this limitation, we introduce VarGes, a method for generating diverse 3D human gestures from speech clips. The intuition of this work lies in enhancing the diversity and naturalness of 3D gesture generation by combining visual and audio information. By integrating these two modalities, we aim to capture the richness of human expressiveness in a way that mirrors how people naturally use both sight and sound to interpret gestures. 
Specifically, in the VarGes, visual information is represented by a style-reference video and encoded as a style code to guide the generation process. In particular, we consider style as a stable and personalized feature that guides generation by influencing the overall characteristics of the action (such as amplitude, rhythm, and force) rather than directly determining the specific action path. With this style guidance, even if the generated actions are different from the style-reference video, they can still be consistent in overall characteristics, thus achieving natural and diverse 3D gesture generation. For example, just as a conductor uses both the visual cues of a musician's body language and the auditory cues of their instrument to guide an orchestra, our approach seeks to leverage the complementary strengths of visual and audio data. 
Figure~\ref{fig1} illustrates the core idea of our approach, where given an arbitrary audio clip, our method generates natural, varied, and realistic gestures, showcasing the potential to overcome the existing limitations in gesture diversity and fluidity.

To this end, we introduce VarGes, a sophisticated framework comprising three pivotal modules: the Variation-Enhanced Feature Extraction (VEFE) Module, the Variation-Compensation Style Encoder (VCSE), and the Variation-Driven Gesture Predictor (VDGP). The VEFE Module is designed to enrich speech-derived features by seamlessly integrating StyleCLIPS extracted from {style-reference} videos. This integration captures information such as gesture rhythm and amplitude, thereby augmenting the stylistic diversity of the input. Following this, the VCSE employs a transformer-based encoder equipped with an additive self-attention pooling layer, enabling the robust encoding of StyleCLIPS into deep learning representations and thereby amplifying their influence on gesture generation. Finally, the VDGP module adeptly combines style codes and MFCC audio features through cross-attention and a cross-conditional autoregressive model, facilitating the generation of diverse and natural 3D gestures. Our experimental results demonstrate the remarkable efficacy of this approach, significantly outperforming existing methods in both gesture diversity and naturalness.

To summarize, the main contributions of our works are as follows:
\begin{itemize}\item We introduce VarGes, a pioneering framework for 3D gesture generation, which integrates speech features with stylistic cues from {style-reference} videos, facilitating the creation of diverse and lifelike 3D human gestures.
\item We establish the Variation-Enhanced Feature Extraction (VEFE) module to extract stylistic clips from 3D human keypoints in {style-reference} videos, and the Variation-Compensation Style Encoder (VCSE) module to encode these clips into deep learning representations, thereby augmenting gesture diversity and expressiveness.
\item We design the Variation-Driven Gesture Prediction (VDGP) module, employing a cross-attention mechanism to merge speech features with style codes and a cross-conditional autoregressive model to govern gesture generation, yielding gestures of enhanced naturalness and variability.
\item Through rigorous quantitative and qualitative assessments, we demonstrate that VarGes significantly outperforms existing approaches, offering a comprehensive array of realistic, voice-synchronized, and high-quality 3D human gestures.
\end{itemize}
\section{Related Works}
\subsection{Extraction of Parametric Data from Videos}
The reconstruction of parametric human shapes is integral to the precise modeling of 3D human bodies. This process involves the extraction of salient features from extensive human body datasets, which are subsequently parameterized into low-dimensional vectors. These parameters facilitate the manipulation and generation of diverse human body shapes, thereby ensuring accurate 3D reconstructions. This methodology provides an efficient and precise framework for representing and reconstructing 3D human forms, offering wide applicability across numerous domains.

Recent advancements in this domain have markedly improved the fidelity and versatility of 3D models. The Skinned Multi-Person Linear (SMPL) model is recognized as a foundational approach, encapsulating human body shape and pose through a set of parameters \cite{42}. Building upon this, the SMPL-X model extends the SMPL framework by incorporating additional shape and pose parameters, thereby enhancing the model's expressiveness and adaptability \cite{43}. The SMPLify-X method further refines this framework by integrating SMPL-X with optimization algorithms, facilitating more precise estimations of human pose and shape from images \cite{43}. Further pushing the envelope, the PyMAF-X method employs multi-task learning, merging parametric models with deep learning to optimize full-body human mesh reconstruction by concurrently addressing multiple related tasks \cite{44}. The Shape and Expression Optimization for 3D Human Bodies (SHOW) \cite{11} further extends these capabilities by optimizing gesture and expression parameters, thereby achieving more realistic and lifelike reconstructions. Nonetheless, these approaches predominantly focus on static body and gesture enhancement, without fully addressing the dynamic generation of gestures that are diverse and contextually aligned with audio inputs.

In our work, we improve the SHOW methodology by incorporating style-reference videos to enhance the synchrony between generated gestures and accompanying audio. By adapting the SHOW framework to process single-speaker videos and utilizing a 3D human keypoint estimation network for extracting keypoints from these references, we introduce additional StyleCLIPS. This augmentation substantially enriches the diversity and realism of the generated 3D human gestures, thereby enhancing their coherence with the input audio.

\subsection{Speech-to-Gesture Generation}
The generation of human gestures from input audio constitutes a multifaceted research domain, synthesizing advancements from speech processing, computer vision, and machine learning. Initial methodologies predominantly relied upon rule-based systems \cite{22}, deploying predefined heuristics to associate gestures with specific vocal inputs. While these foundational approaches established a basis, they frequently lacked the adaptive capacity to encapsulate the nuanced complexities of human gestural expression. In response, statistical models emerged \cite{23}, designed to capture the intrinsic variability and sophistication of gestures. These models \cite{24} endeavored to learn individual speaker styles through probabilistic representations, employing hidden Markov models (HMMs) \cite{25} to harness prosodic speech features for gesture prediction. Additionally, statistical frameworks were integral to synchronizing speech with gestures in embodied conversational agents (ECAs) \cite{26}.

The advent of deep learning has precipitated a paradigm shift in the field of human gesture generation. The proliferation of deep learning techniques has obviated the necessity for manual gesture lexicons and mapping rules, fostering a renaissance in voice-driven gesture synthesis. Contemporary methodologies leverage a panoply of techniques, including recurrent neural networks (RNNs) \cite{27,28,29}, generative adversarial networks (GANs) \cite{30,31,32}, and diffusion models \cite{12,33,34,35}, to refine the synthesis of human gestures. Furthermore, autoencoder architectures such as variational autoencoders (VAEs) \cite{10,36,37,38}, vector quantized variational autoencoders (VQ-VAEs) \cite{11,39,40}, and hybrid models integrating flows with VAEs \cite{41} have been explored to engender diverse gestural outputs. Our approach builds upon these advancements by integrating VQ-VAEs with cross-conditioned autoregressive models to proficiently map speech to both hand and body gestures.

Despite these advancements, extant methodologies frequently encounter limitations regarding gesture diversity, often attributable to simplistic identity labels that confine the range of generated gestures within the dataset constraints. For example, prior investigations such as \cite{11} utilized video data from a limited cohort to infer 3D human poses from speech, and \cite{13} employed CNN and GAN architectures with data from ten individuals to map speech signals to gestures. While these approaches exhibit innovation, they often fall short in capturing extensive gestural variability. Our method transcends these limitations by introducing supplementary stylistic influences through StyleCLIPS, which are further encoded into style codes. These style codes guide the generation process by shaping overall gesture characteristics, such as amplitude, rhythm, and intensity, rather than prescribing specific gesture trajectories. This approach enhances gestural diversity. By employing a cross-attention mechanism to integrate audio features with style codes, our methodology significantly improves both the diversity and realism of the generated 3D gestures.

\section {Method}
\setcounter{subsection}{0}
\subsection{Overview and Problem Formulation}
Our approach is designed to enhance the modulation of generated 3D full-body human gestures, with a focus on augmenting both their diversity and naturalness. The framework processes voice audio input \( A= \left \{a_1,...,a_N \right\} \), where \( N \) represents the total number of frames, to produce a corresponding sequence of full-body gestures \( G= \left \{g_1,...,g_N \right\} \), with each \( g_i \) representing the human full-body gesture at frame \( i \). To achieve gestures that are contextually apt and varied, the system integrates additional modalities such as style-reference videos and speaker identity, {which contribute to shaping the overall features for gesture generation, thereby enriching the diversity and naturalness of the gesture synthesis process.} The primary objective is to optimize the modulation of these gestures, achieving a harmonious balance between variability and natural appearance. The overarching aim of our method is formalized as follows.
\begin{center}
\begin{equation}
    \arg\min_{\hat{G} } \left \| G-\hat{G} \left ( A,\left \{ g_1,...,g_N \right \}  \right )  \right \| 
\end{equation}
\end{center}

where \( \left \{ g_1,...,g_N \right \} \) denotes the initial pose sequence. This formulation seeks to minimize the discrepancy between the target gesture sequence \( G \) and the synthesized sequence \( \hat{G} \), thereby ensuring the generated gestures exhibit both diversity and authenticity.

\begin{figure*}[h!t]
    \centering
    \includegraphics[width=1\linewidth]{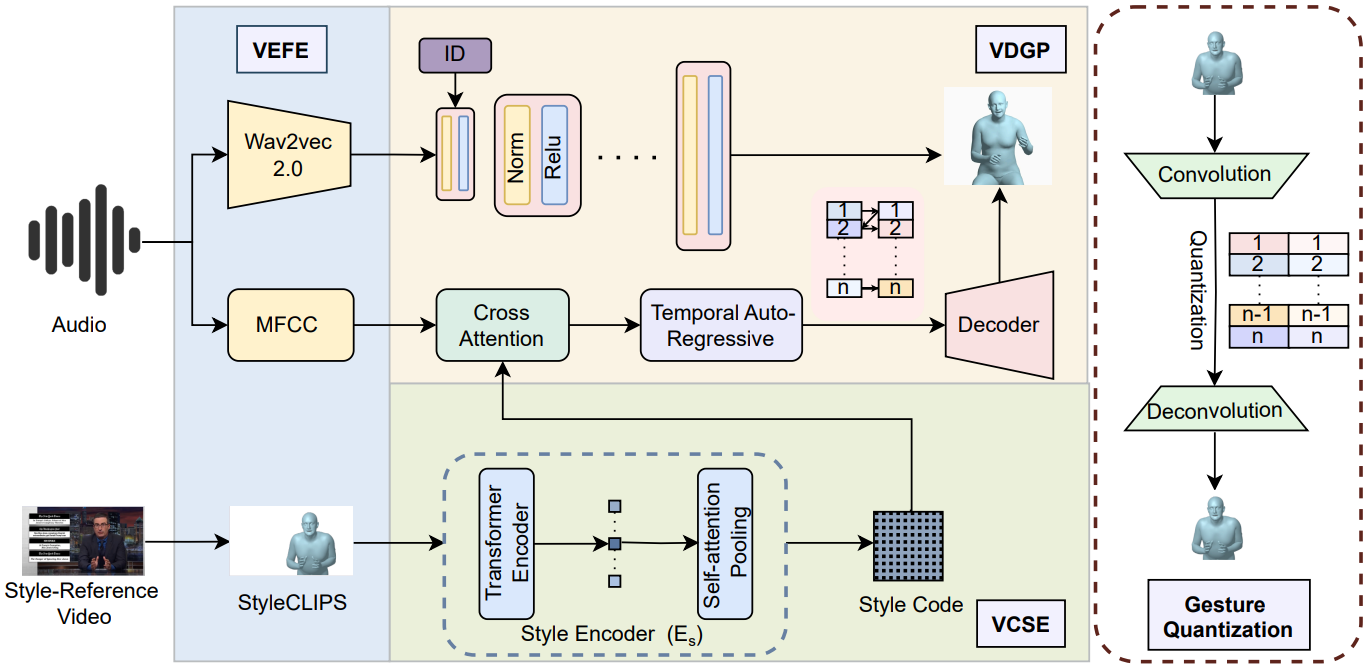}
    \caption{Overview of the VarGes framework. VarGes comprises three modules: The Variation-Enhanced Feature Extraction (VEFE) module extracts key features from speech using Wav2vec 2.0 and MFCC, filtering noise with StyleCLIPS from style-reference videos. The Variation-Compensation Style Encoder (VCSE) module encodes style-clips into deep feature style codes with a transformer-based encoder and self-attention pooling. The Variation-Driven Gesture Predictor (VDGP) module fuses style codes and MFCC through cross-attention and a temporal autoregressive network, incorporating identity information to boost gesture diversity and naturalness. Action quantization is applied during training to further increase action variability.}
    \label{Overview2}
\end{figure*}

The overall methodology is illustrated in Figure~\ref{Overview2}, which provides a high-level overview of the framework's components. The process begins with the extraction of audio features through Wav2vec 2.0, supplemented by style clips derived from style-reference videos. These inputs feed into the Variation-Enhanced Feature Extraction (VEFE) module, where features are processed and combined via a cross-attention mechanism, incorporating identity features to enhance gesture variation. The Variation-Compensation Style Encoder (VCSE) module encodes the stylistic information using a transformer encoder and self-attention pooling, generating a style code that is integrated into the temporal auto-regressive model within the Variation-Driven Gesture Predictor (VDGP) module. The final gesture sequence is predicted by this model and refined by gesture quantization using VQ-VAE, ensuring the generation of high-quality, realistic gestures.

In this process, our approach relies on two key components: Gesture Representation and Gesture Quantization. The representation of full-body gestures is achieved through a comprehensive character model, which includes 300 dimensions for full-body shape, 156 dimensions for full-body gestures (comprising 3 dimensions for chin posture, 63 dimensions for body posture, and 90 dimensions for hand posture), 3 dimensions for camera pose, 3 dimensions for translation, and 100 dimensions for facial expressions. Specifically, hand gestures at frame $i$ are represented by \(\ g_{i}^{H} \in R^{90}\), while body gestures are represented by \(g_{i}^{B} \in R^{63}\ \). A sequence of hand gestures is collectively denoted as \(G^{H} = \ \{ g_{1}^{H},...,g_{N}^{H}\}\), and a sequence of body gestures as \(G^{B} = \ \{ g_{1}^{B},...,g_{N}^{B}\}\), where $N$ represents the number of frames.

To effectively encode and generate these gestures, we employ a dedicated VQ-VAE (Vector Quantized Variational Autoencoder) designed specifically for hand and body gesture quantization. During the training phase, given the hand gesture sequence \(G^{H} \in R^{N \times 90}\) and the body gesture sequence \(G^{B} \in R^{N \times 63}\), where $N$ denotes the number of frames, we first use a temporal convolutional network to jointly encode the sequences \(G^{H}\) and \(G^{B}\) into feature sequences \(\ g^{H} \in R^{N^{'} \times C}\) and {
\(g^{B} \in R^{N^{'} \times C}\)}, respectively. Here, \(N^{'} = \frac{N}{d}\) represents the down-sampled temporal length, and $C$ is the feature channel dimension. This encoding process can be expressed as $g = E(G)$. To capture the variations in gestures effectively, the encoded features \(g_{i}^{H}\) and \(g_{i}^{B}\) are quantized by mapping each feature to its closest codebook element Z\textsuperscript{H} and Z\textsuperscript{B}, as follows:

\begin{align}
    \begin{split}
    z_{i}^{H} = Q(g^{H}) = \ \arg\ \min_{z_{m}^{H} \in Z^{H}}\left\| g_{i}^{H} - z_{m}^{H} \right\| \\
    z_{i}^{B} = Q(g^{B}) = \ \arg\ \min_{z_{m}^{B} \in Z^{B}}\left\| g_{i}^{B} - z_{m}^{B} \right\|
    \end{split}
    \label{eq2}
\end{align}

Finally, the decoder takes the codebook elements \(Z^{H}\) and  \(Z^{B}\)
projects back to the motion space as a pose sequence , which can be
formulated as:

\begin{align}
\begin{split}
G^{H} = \ D(Z^{H})\  = D(Q(E(G^{H})))  
\\
G^{B} = \ D(Z^{B})\  = D(Q(E(G^{B}))) \end{split}\label{eq3}
\end{align}

Thus the encoder, decoder and codebook can be trained by optimizing the following objective:
\begin{align}
L_{VQ\_ VAE} = L_{rec}(\widehat{G},G) + \left\| sg\lbrack g\rbrack - z \right\| + \beta\left\| g - sg\lbrack z\rbrack \right\| \label{eq4}
\end{align}

Where \(L_{rec}\) is the reconstruction loss, $sg$ denotes the stop-gradient operation, and the term
\(\left\| g - sg\lbrack z\rbrack \right\|\) represents the 'commitment loss' with a weighting factor $\beta $ \cite{42}.

\subsection{Variation-Enhanced Feature Extraction Module}
Traditional methods for 3D gesture generation often rely on speech feature extraction, leading to limited variation due to the constraints of audio data. To overcome this, we propose the Variation-Enhanced Feature Extraction (VEFE) Module, which combines advanced speech features from Wav2vec 2.0 and MFCC with stylistic cues from style-reference videos via StyleCLIPS, resulting in more diverse and contextually appropriate gestures.

\textbf{Wav2vec 2.0 Encoder.} Given the strong correlation between speech and facial animation, the audio encoder $E_A$ is designed to extract high-level speech features. We utilize the state-of-the-art, self-supervised Wav2vec 2.0 model \cite{45} to capture rich phoneme information. The input audio is encoded into latent features through a multi-layer convolutional network, partially masked, and processed by a transformer to generate a 768-dimensional speech feature representation. A linear projection layer then reduces the feature dimension to 256.

\textbf{MFCC Representation.} To extract articulation-related information from the audio while filtering out articulation-irrelevant components such as phonemes that might influence hand and body gestures, we utilize Mel Frequency Cepstral Coefficients (MFCC). The audio signals are represented as \(A^{M} \in R^{64 \times N}\), effectively isolating the relevant features for gesture generation.

\textbf{StyleCLIPS.} Character gesture variation is primarily determined by dynamic patterns of the head, hands, and body, which are independent of extraneous factors such as clothing, hairstyle, and lighting in the style-reference video. To minimize distractions from these irrelevant elements, we convert the style-reference video into sequential gesture parameters \(G\ \epsilon\ R^{156 \times N}\), termed as StyleCLIPS. StyleCLIPS encapsulate overall characteristics with a personalized style, such as gesture rhythm and amplitude, which contribute to the unique stylistic features of the generated gestures. In both the training and inference phases, the style-reference video serves as an input, providing supplemental stylistic information for the gesture generation process. Notably, the style-reference video can be arbitrary and does not need to align with the input audio. Initially, DECA \cite{49}, PIXIE \cite{50}, and PyMAF-X \cite{51} are used to set up parameters for facial expression, jaw, body, and hand movements. These parameters are then optimized through a module that integrates human contours from DeepLab V3 \cite{52}, facial landmarks from MediaPipe, and facial shapes from MICA to ensure accurate contouring. This process ensures that the rendered SMPL-X body remains within the human mask, while photometric loss is used to capture detailed facial features. The result is a highly realistic 3D full-body mesh synchronized with audio, enhancing the accuracy of full-body reconstruction by optimizing posture and expression.
\subsection{Variation-Compensation Style Encoder Module}
As described in VEFE, we extract StyleCLIPS from the style-reference video. However, simply encoding these StyleCLIPS may not fully capture their complex features, potentially leading to the loss of critical information. To address this, we introduce the Variation-Compensation Style Encoder (VCSE) Module, designed to comprehensively encode and extract style features. The style encoder \(E_{s}\) converts StyleCLIPS into a deep learning representation known as the style code. To effectively model the dynamic hand and body posture patterns, we developed a transformer-based style encoder enhanced with an additional self-attention pooling layer.

The process begins with the 3D SMPL-X pose parameters as input. These parameters are adjusted to the desired dimensions through a linear layer before position encoding is applied. The style encoder treats the processed sequential 3D SMPL-X pose parameters as input tokens. Since the pose style within a sequence can often be identified by a few key frames, irrelevant or padded sections should be excluded from consideration. To handle variable-length sequences, the encoder utilizes a fill mask that ensures only relevant information is processed. Specifically, a feedforward neural network, coupled with the self-attention pooling layer, weighs the attention of each region after segmenting the input pose sequence. The attention weight assigned to each region reflects the contribution of each frame to the overall style of the sequence. After modeling the temporal correlations among the tokens, the resulting style vectors are multiplied by attention weights in the self-attention pooling layer to produce the final style code \(s \in R^{d_{s}}\):
\begin{align}
s = softmax(W_{s}S)S^{T}\
\label{eq5}
\end{align}
Where \(W_{s} \in R^{1 \times d_{s}}\) is a trainable parameter,
\(S = (s_{1},...,s_{n}) \in R^{d_{s} \times N}\) is a sequence of style
codes obtained by the style encoder, and \(d_{s}\) is the dimension of each style vector.

\begin{figure}[h!t]
    \centering
    \includegraphics[width=.9\linewidth]{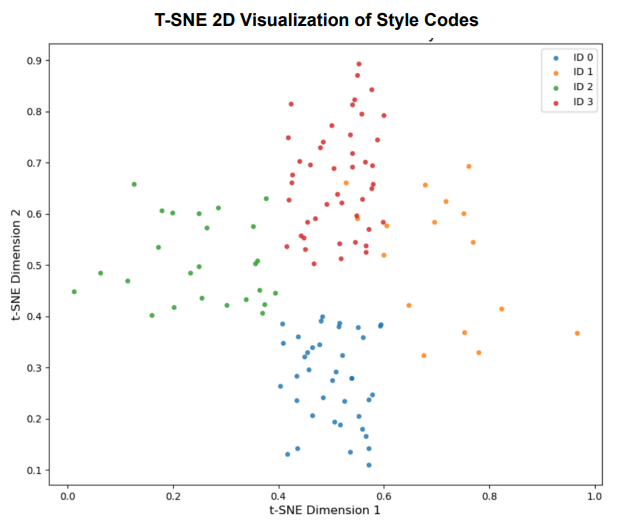}
    \caption{{t-SNE visualization of style code Distribution. This figure illustrates the t-SNE visualization of the Style Code learned from videos associated with four different IDs.}}
    \label{tsne}
\end{figure}

{To evaluate the learned style code, we employ t-SNE \cite{55} (t-distributed Stochastic Neighbor Embedding), a widely used dimensionality reduction technique, to visualize the distribution of style codes extracted from style-reference videos. Focusing on the test set of the SHOW dataset, which includes videos from four distinct individuals, we demonstrate how the style codes cluster based on their stylistic IDs. By reducing the style codes to two dimensions using t-SNE with a perplexity of 30 and a learning rate of 200, we achieve clear clustering of the style codes, as shown in Figure~\ref{tsne}. Each point in the visualization represents a style code from a reference video, color-coded by its stylistic ID. The results reveal a distinct separation between style codes corresponding to different stylistic IDs, confirming that the proposed Style Encoder effectively captures and differentiates stylistic features. This clustering not only validates the ability of the style code to encode style-specific characteristics but also highlights its robustness in handling diverse reference videos.}

\subsection{Variation-Driven Gesture Predictor Module}
Traditional methods for gesture generation from audio features typically rely on regression techniques, which often only capture basic rhythms and amplitude changes, leading to gestures that lack variation and naturalness. To address this, we propose the Variation-Driven Gesture Predictor (VDGP) Module. This module employs a cross-attention mechanism to facilitate high-level interaction between different modal features. The fused features are then input into a temporal auto-regressive model, utilizing codebook vectors to predict gestures. This approach not only captures fundamental rhythms and variations in speech but also integrates complex gesture styles and details from style-reference videos, enhancing the diversity and naturalness of the generated gestures.

The Cross Attention layer modulates the MFCC features \(A^{M}\) using the style code $s$. The Key and Value matrices are derived from the style code $s$, while the Query matrix comes from the MFCC \(A^{M}\). Specifically, the inputs \(A^{M}\) and $s$ are projected to obtain the Query \(Q_{A}\), Key matrix \(K_{s}\), and Value
\(V_{s}\):
\begin{align}
Q_{A} = AW_{Q}\nonumber 
\\
K_{s} = sW_{K}\label{eq6}
\\
V_{s} = sW_{V}\nonumber 
\end{align}
The attention mechanism is then applied as:
\begin{align}
F = Attention(Q_{A},K_{s},V_{s}) = softmax(\frac{Q_{A}K_{s}^{T}}{\sqrt{d_{k}}})V_{s} \
\label{eq7}
\end{align}
Where \({\ d}_{k}\) represents the dimension of the key, value, and query sets. The resulting attention score matrix reflects the degree of correlation between the two modalities, with the product of this matrix and the Value matrix \(V_{s}\) representing the adaptation of the style code to the MFCC features. The fused features $F$ are then passed into a temporal auto-regressive model, which generates a sequence of codebook vector indices $X^{H}$ and $X^{B}$. The model uses mutual information to predict the current hand gesture and body pose code indices based on past gestures and poses. This process can be formalized as:
\begin{align}
p({X_{1:N}^{B}} , {X_{1:N}^{H}} | {A_{1:N}}, F) &= \Pi_{i = 1}^{N'}{p({x_{i}^{B}}|{x_{< i}^{B}} , {x_{< i}^{H}} , {a_{\leq i}} , F)} \nonumber \\
&\quad \times p({x_{i}^{H}}|{x_{< i}^{H}} , {x_{< i}^{H}} , {a_{\leq i}} , F)
\label{eq8}
\end{align}
Finally, the generated indices \(X^{H}\) and \(X^{B}\) are used to etrieve quantized motion elements from the learned codebooks \(Z^{H}\) and \(Z^{B}\). These elements are then decoded by the VQ-VAE to produce the final hand gestures \(\widehat{G^{H}}\) and body poses \(\widehat{G^{B}}\).
\section {Experiments}
\setcounter{subsection}{0} 
\begin{table*}[h!t]
\center
\caption{Quantitative comparison on the SHOW dataset. A comparison of our method (VarGes) against state-of-the-art approaches using three key metrics: Variation, FGD, and BC. Higher Variation indicates greater diversity in the generated gestures, lower FGD reflects closer alignment with ground truth gestures, and a BC score closer to the ground truth value indicates better synchronization between gestures and audio.}
\label{Comparison_quantitative}

\renewcommand\arraystretch{1.2} 
\setlength{\tabcolsep}{5pt} 

\begin{tabular*}{\linewidth}{@{\extracolsep{\fill}} llll } 
\toprule
\multicolumn{1}{c}{Method}    & Variation
$ \uparrow $
      & FGD$ \downarrow $           & BC (GT 0.8680)  \\ \hline
{GT}       & {1.0069}            & {0}        & {0.8680}           \\
Audio2Gesture\cite{10}        & 0.24            & 203.990        & 0.943           \\
LS3DCG(pretrained)\cite{13}   & 0               & 239.170        & 0.9476          \\
LS3DCG(re-train)\cite{13}     & 0               & 245.733        & 0.9370          \\
TalkSHOW(pretrained)\cite{11} & 0.8796          & 70.215         & 0.8721          \\
TalkSHOW(re-train)\cite{11}   & 0.9265          & 46.356         & 0.8762          \\
VarGes                        & \textbf{0.9977} & \textbf{5.463} & \textbf{0.8690} \\ 
\bottomrule
\end{tabular*}
\end{table*}

\subsection{Implementation and Training Details}
Our implementation follows the data preparation protocol established in TalkSHOW \cite{11}. The dataset is randomly shuffled and partitioned into training, validation, and test sets in an 8:1:1 ratio. For model optimization, we use the Adam optimizer with parameters $\beta _{1} = 0.9$, $\beta _{2} = 0.999$ and a learning rate of 0.0001. The commitment loss weight is set to 0.25. The Variation-Compensation Style Encoder (VCSE) module , which utilizes a transformer style encoder with an additional self-attention pooling layer, and the Variation-Driven Gesture Predictor (VDGP) module , which employs a cross-attention mechanism, are trained with a batch size of 128 and a sequence length of 88 frames over 100 epochs. The model is implemented using the PyTorch framework and trained on a single NVIDIA GeForce RTX 3090 GPU for approximately five days.

\subsection{Dataset}
To evaluate and benchmark our approach against existing methods in audio-to-gesture generation, we utilize the SHOW dataset \cite{11}, a high-quality resource specifically designed for this task. The SHOW dataset comprises synchronized speech audio and 3D full-body mesh data for four distinct speakers. These meshes are reconstructed using SMPL-X \cite{43} parameters from video recordings captured at 30 frames per second, while the corresponding audio is sampled at 22 kHz. For robust evaluation, the dataset is partitioned into training, validation, and test sets with a distribution of 80\%, 10\%, and 10\%, respectively, ensuring comprehensive coverage of various gesture styles and speech patterns.

\subsection{Evaluation Metrics}
To rigorously assess the performance of our proposed VarGes framework, we employ the following evaluation metrics:

\textbf{Variation} quantifies the diversity of the generated motion sequences. Following the approach used in \cite{46}, we calculate the variance across 16 samples to measure the diversity of the generated gestures. Specifically, variation is computed as:

\begin{align}
Variation=\frac{1}{N} \sum\limits_{i=1}^{N} \left \| (Var \hat{g_{i}}) \right \| _{2} 
\label{eq9}
\end{align}
where $\hat{g_{i}}$ denotes the $i-th$ generated gesture sample and $Var$ represents the variance operation. This metric provides insights into the range of motion styles produced by our framework.

\textbf{Frechet Gesture Distance (FGD)} measures the realism of the generated gestures by evaluating the distribution distance between the ground truth gestures and the synthesized ones \cite{15}. This is achieved by comparing the feature distributions encoded by a pre-trained encoder. The FGD is computed as:
\begin{align}
FGD(G,\hat{G} ) = \left \| \mu _{r} - \mu _{g}   \right \| ^{2} + Tr( {\textstyle \sum\limits_{r}} +  {\textstyle \sum\limits_{g}} -2({\textstyle \sum\limits_{r}{\textstyle \sum\limits_{g}}})^{1/2} )
\label{eq10}
\end{align}

wher $\mu _{r}$ and $\textstyle \sum_{r}$ are the mean and covariance of the latent feature distribution $Z_{r}$ of real human gestures $G$, and $\mu _{g}$ and ${\textstyle \sum_{g}}$ are those of the generated gestures $\hat{G}$. This metric assesses how closely the generated gestures align with real human motion patterns.

\textbf{Beat Consistency Score (BC)} evaluates the synchronization between generated gestures and their corresponding audio by measuring the alignment between audio beats and motion beats \cite{32}. The BC score is calculated as:
\begin{align}
BC = \frac{1} {G} \sum_{b_{G}\in G } \exp \left( -\frac{\min_{b_A \in A} \left( \left\| b_G - b_A \right\|^2 \right)}{2\sigma^2} \right)
\label{eq11}
\end{align}

where $G$ and $A$ denote the sets of kinematic and audio beats respectively, and $\sigma$ is a normalization parameter empirically set to 0.1. This metric assesses how well the motion beats align with the beats in the audio, reflecting the temporal synchronization between gestures and speech.

\subsection{Quantitative Evaluation}
This section provides a thorough quantitative assessment of the VarGes framework. We first compare its performance against state-of-the-art methods using key metrics such as Variation, FGD, and BC. Following this, an ablation study explores the impact of specific components within the VarGes architecture, particularly {the influence of style-reference videos,} the style encoder and the integration methods for style code and MFCC.
\subsubsection{Comparisons with state-of-the-art methods}
We evaluate the performance of our VarGes framework against several state-of-the-art methods, including Audio2gestures (A2G) \cite{10}, LS3DCG \cite{13}, and TalkSHOW \cite{11}. {Additionally, we include GT (Ground Truth) values as a benchmark for these metrics, offering a clearer context for evaluating the performance of generated gestures against real-world motion data.} Audio2gestures uses a Variational Autoencoder (VAE) to separate gesture latent space into shared and audio-independent codes, enabling diverse gesture generation. LS3DCG leverages a CNN and GAN-based architecture to exploit the correlation between facial expressions and gestures, providing a method for generating 3D body motions from in-the-wild speaker videos. TalkSHOW introduces a high-quality dataset of 3D full-body meshes synchronized with speech, using an encoder-decoder for face modeling and a VQ-VAE-based approach for body and hand gestures, resulting in realistic gesture outputs.
\begin{table*}[t]
\centering 
\caption{Ablation study of style encoder configurations. A comparative analysis of different style encoder configurations within our framework, evaluating their impact on three key metrics. Higher Variation indicates greater diversity in the generated gestures, lower FGD reflects closer alignment with ground truth gestures, and a BC score closer to the ground truth value indicates better synchronization between gestures and audio.}
\label{Ablation_configurations}
\renewcommand\arraystretch{1.2} 
\setlength{\tabcolsep}{5pt} 
\begin{tabular*}{\linewidth}{@{\extracolsep{\fill}} llll }
\toprule
\multicolumn{1}{c}{Configuration}                  & \multicolumn{1}{c}{Variation ($\uparrow$)}  & \multicolumn{1}{c}{FGD ($\downarrow$)}           & \multicolumn{1}{c}{BC (GT 0.8680)}  \\
\midrule
{GT}       & {1.0069}            & {0}        & {0.8680}           \\
CNN-based style encoder                             & 0.8928          & 104.988        & 0.8719          \\
FCN-based style encoder                             & 0.9341          & 97.004         & 0.8702          \\
Transformer encoder layers=6 (no polling)           & 0.9026          & 86.151         & 0.8691          \\
Transformer encoder layers=6 (Average polling)      & 0.9413          & 22.250         & 0.8693          \\
Transformer encoder layers=6 (SelfAttentionPooling) & 0.9698          & 17.127         & 0.8698          \\
Transformer encoder layers=4 (SelfAttentionPooling) & 0.8978          & 14.407         & 0.8695          \\
Transformer encoder layers=7 (SelfAttentionPooling) & 0.9268          & 108.451        & 0.8695          \\
Transformer encoder layers=8 (SelfAttentionPooling) & \textbf{0.9731} & \textbf{8.072} & \textbf{0.8677} \\
Transformer encoder layers=9 (SelfAttentionPooling) & 0.9239          & 170.754        & 0.8697          \\
Transformer encoder layers=8 (no polling)           & 0.9008          & 46.819         & 0.8707          \\
Transformer encoder layers=8 (Average polling)      & 0.9054          & 63.872         & 0.8693          \\ 
\bottomrule
\end{tabular*}
\end{table*}

\begin{table*}[h!t]
\center
\caption{Ablation study on style code and MFCC integration. Comparison of two integration methods: direct injection (w/o cross-attention) and cross-attention. Higher Variation values indicate greater gesture diversity, lower FGD values reflect better alignment with ground truth, and BC scores closer to the ground truth indicate improved synchronization between gestures and audio.}
\label{Ablation_MFCC}
\renewcommand\arraystretch{1.2} 
\setlength{\tabcolsep}{5pt} 
\begin{tabular*}{\linewidth}{@{\extracolsep{\fill}} llll }
\toprule
                                       & Variation
$ \uparrow $
      & FGD$ \downarrow $           & BC (GT 0.8680)  \\ \hline
{GT}       & {1.0069}            & {0}        & {0.8680}           \\
w/o cross-attention (Direct Injection) & 0.9731          & 8.072          & \textbf{0.8677} \\
Cross Attention                        & \textbf{0.9977} & \textbf{5.463} & 0.8690          \\ 
\bottomrule
\end{tabular*}
\end{table*}

Our quantitative results, as shown in Table~\ref{Comparison_quantitative}, demonstrate that VarGes outperforms these baseline methods across all key metrics. VarGes achieves a Variation score of 0.9977, significantly higher than the closest competitor, TalkSHOW (re-trained), at 0.9265, indicating superior gesture diversity. Furthermore, VarGes records an FGD of 5.463, substantially lower than all other methods, highlighting the enhanced realism of our generated gestures. In terms of BC, VarGes achieves 0.8690, closely matching the ground truth of 0.8680, showcasing better synchronization between gestures and audio than the baseline methods. These results confirm that VarGes not only generates more diverse and realistic gestures but also ensures better alignment with the input audio.

\subsubsection{Ablation Study}
{To further evaluate the contributions of individual components in our framework and gain deeper insights into their roles in gesture generation, we conduct a series of ablation studies.}

\begin{figure*}
    \centering
    \includegraphics[width=1\linewidth]{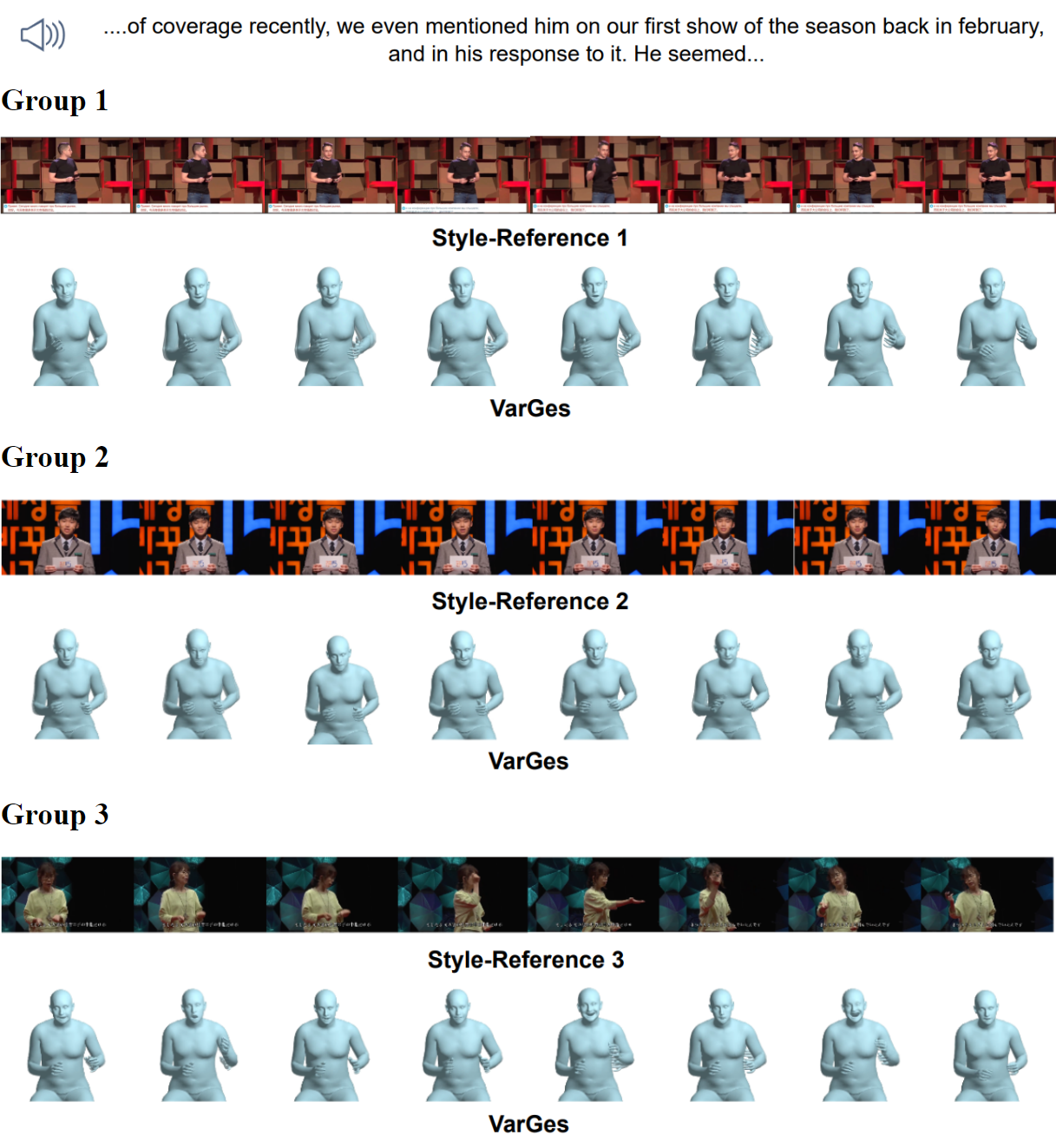}
    \caption{{Visualization of the same audio with different reference-style videos. The figure illustrates the gesture generation results of our model when provided with identical audio input and different style-reference videos. The generated gestures exhibit synchronization with the audio while adapting to the distinct stylistic characteristics of each reference video, demonstrating the model's ability to achieve both diversity and naturalness in gesture generation.}}
    \label{same}
\end{figure*}

\textbf{{Effect of Style-Reference Videos on Generated Gesture Variation.}} {To explore the effect of style-reference videos on the variation and diversity of generated gestures, we perform an ablation study by analyzing the impact of using different style-reference videos with the same audio input. This study investigates whether the model can generate stylistically diverse gestures while maintaining synchronization with the audio. The results are visualized in Figure~\ref{same}, the first group's style-reference video features a few raised hand gestures with mostly horizontal movements.  The corresponding generated gestures align with this style, adding subtle variations in the fingers for increased realism and diversity.In the second group, the style-reference video features consistently horizontal gestures with both hands holding a card, and minimal finger movement. The generated gestures similarly lack upward motion, but include slight finger variations for subtle dynamics. In the third group, the gestures in the style-reference video feature multiple arm-raising movements. Similarly, the corresponding video generated by our model includes several arm-raising actions, complemented by noticeable finger variations that enhance the naturalness and expressiveness of the gestures. In summary, these results illustrate that the inclusion of StyleCLIPS enhances, rather than limits, the diversity and expressiveness of the generated gestures.} 

\textbf{{Effectiveness of Transformer Style Encoder with Self-Attention Pooling.}} To validate the effectiveness of our proposed transformer style encoder with an additional self-attention mechanism pooling layer, we conduct a comprehensive ablation study to evaluate the impact of different components of the style encoder on the key metrics. Various configurations are tested, including different encoder types (CNN-based, FCN-based) and the influence of pooling mechanisms within transformer layers. The results, presented in Table~\ref{Ablation_configurations}, clearly indicate that the use of transformer encoder layers with self-attention pooling consistently outperforms other configurations. Specifically, the configuration with 8 transformer encoder layers combined with self-attention pooling achieves the highest Variation score and the lowest FGD, demonstrating a significant improvement in gesture diversity and realism. This suggests that incorporating self-attention pooling into the transformer encoder layers is critical for enhancing the variation and quality of the generated gestures.

\textbf{{Impact of Variation Enhancement Module with Cross-Attention Mechanism.}} To assess the impact of the variation enhancement module based on a cross-attention mechanism, we conduct an additional ablation study. This study compared two configurations: one utilizing cross-attention mechanisms for feature integration and another directly injecting the style code and MFCC into the model. The results, summarized in Table~\ref{Ablation_MFCC}, demonstrate that the cross-attention mechanism significantly enhances both the Variation and FGD metrics compared to the direct injection method. However, the BC score of the cross-attention configuration is slightly lower due to jitter or abnormal motion that artificially boosts the BC score. Nonetheless, the cross-attention mechanism is shown to be effective in capturing the nuances required to generate diverse and context-appropriate gestures.

\begin{table*}[t]
\centering
\renewcommand{\arraystretch}{1.2} 
\setlength{\tabcolsep}{8pt}       
\caption{User study evaluation items and corresponding Cronbach's $\alpha$ values.}
\label{Userstudy_item}
\begin{tabular}{p{4cm} p{8cm} c} 
    \toprule
    Scale & Item & Cronbach's $\alpha$ \\
    \midrule
    \multirow{3}{*}{Human Similarity} 
        & Gestures resembled human movements. & \multirow{3}{*}{0.966} \\
        & Gestures were lifelike. & \\
        & Gestures appeared natural for a human. & \\
    \midrule
    \multirow{3}{*}{Speech-Gesture Correlation} 
        & Gestures were well synchronized with the speech. & \multirow{3}{*}{0.963} \\
        & Gestures matched the rhythm of the speech. & \\
        & Gestures matched intonation of speech. & \\
    \midrule
    \multirow{3}{*}{Gesture Smoothness} 
        & Gestures transitioned smoothly. & \multirow{3}{*}{0.969} \\
        & Gestures were fluid. & \\
        & Gestures did not appear jerky or abrupt. & \\
    \midrule
    \multirow{3}{*}{Gesture Naturalness} 
        & Gestures appeared natural. & \multirow{3}{*}{0.969} \\
        & Gestures were realistic. & \\
        & Gestures were appropriate for the context. & \\
    \midrule
    \multirow{3}{*}{Gesture Variation} 
        & Gestures were varied and not repetitive. & \multirow{3}{*}{0.969} \\
        & Gestures showed diversity. & \\
        & Gestures included a range of movements. & \\
    \bottomrule
\end{tabular}
\end{table*}

\begin{table*}[h!t]
\centering
\caption{User study results. Comparison of gesture quality between TalkSHOW and our method, based on a 1-7 rating scale across Human Similarity, Speech-Gesture Correlation, Gesture Smoothness, Gesture Naturalness, and Gesture Variation. Higher scores indicate better performance in the corresponding item and scale. SubM represents the mean of the three indicators for each item, and Mean represents the mean of each indicator as a whole.}
\label{Userstudy_score}

\renewcommand{\arraystretch}{1.2} 
\begin{tabularx}{\linewidth}{lXcccc}
\hline
Scale & Item & TalkSHOW & & \textbf{Ours} & \\
      &      & SubM  & Mean & SubM & Mean \\
\hline
\multirow{3}{*}{Human Similarity} 
    & Gestures resembled human movements. & 4.22 & 4.12 & 5.61 & 5.56 \\
    & Gestures were lifelike. & 4.09 & & 5.52 & \\
    & Gestures appeared natural for a human. & 4.05 & & 5.55 & \\
\hline
\multirow{3}{*}{Speech-Gesture Correlation} 
    & Gestures were synchronized with the speech. & 4.22 & 4.11 & 5.50 & 5.42 \\
    & Gestures matched the rhythm of the speech. & 4.07 & & 5.31 & \\
    & Gestures matched speech intonation. & 4.05 & & 5.45 & \\
\hline
\multirow{3}{*}{Gesture Smoothness} 
    & Gestures transitioned smoothly. & 4.17 & 4.08 & 5.72 & 5.61 \\
    & Gestures were fluid. & 4.15 & & 5.56 & \\
    & Gestures did not appear jerky or abrupt. & 3.92 & & 5.56 & \\
\hline
\multirow{3}{*}{Gesture Naturalness} 
    & Gestures appeared natural. & 3.94 & 3.91 & 5.66 & 5.56 \\
    & Gestures were realistic. & 4.00 & & 5.55 & \\
    & Gestures were appropriate for the context. & 3.81 & & 5.48 & \\
\hline
\multirow{3}{*}{Gesture Variation} 
    & Gestures were varied and not repetitive. & 4.20 & 4.13 & 5.53 & 5.40 \\
    & Gestures showed diversity. & 4.15 & & 5.35 & \\
    & Gestures included a range of movements. & 4.06 & & 5.31 & \\
\hline
\end{tabularx}
\end{table*}

\subsection{Qualitative Evaluation}
This section evaluates VarGes qualitatively through user studies and visual comparisons, demonstrating its superiority in generating varied, natural, and contextually appropriate gestures compared to state-of-the-art methods.
\subsubsection{User Study Setup and Analysis}
To deepen our understanding of the visual performance of our proposed method, we conduct a user study to evaluate its effectiveness compared to the state-of-the-art approach.

\textbf{Participant Characteristics.} We recruit 31 undergraduate students through an online community, comprising 8 males and 23 females, aged between 18 and 29 years (mean age: 22 years). The participants came from diverse academic backgrounds, including majors such as Communication and Information Systems, Digital Arts, Information Communication Studies, Data Science and Big Data Technology, Intelligent Science and Technology, Mathematics, and Broadcast Television Engineering, spanning a total of 15 different fields. Among them, 18 participants had research experience in deep learning, 14 had some knowledge of the digital human field, and 5 had conducted research specifically in this area.

\textbf{{MOS-Based Evaluation.}} The user study involved 31 participants evaluating 10 groups of 20 videos, each approximately 10 seconds long, generated by our method and other state-of-the-art techniques. The widely recognized Mean Opinion Scores (MOS) \cite{48} rating protocol was employed for this evaluation. Participants were provided with a questionnaire (as detailed in Table~\ref{Userstudy_item}) to assess the videos. They were asked to watch all the synchronized speech and gesture videos and then rate them on five key aspects: (1) Human Similarity; (2) Speech-Gesture Correlation; (3) Gesture Smoothness; (4) Gesture Naturalness; and (5) Gesture Variation. Each aspect was evaluated using three items on a seven-point Likert scale, ranging from 1 (strongly disagree) to 7 (strongly agree). The reliability of the questionnaire was confirmed, with Cronbach’s $\alpha $ values exceeding 0.9 for all scales, as shown in Table~\ref{Userstudy_item}.


\begin{figure*}
    \centering
    \includegraphics[width=1\linewidth]{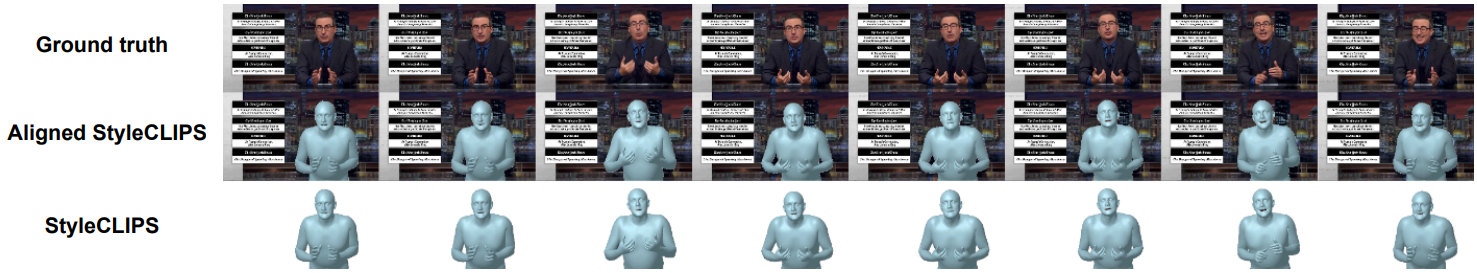}
    \caption{Ground truth comparison and 3D mesh visualization with StyleCLIPS. A side-by-side comparison between the original video frames and the corresponding 3D meshes generated using StyleCLIPS.}
    \label{Styleclips}
\end{figure*}

\begin{figure*}[h!t]
    \centering
    \includegraphics[width=1\linewidth]{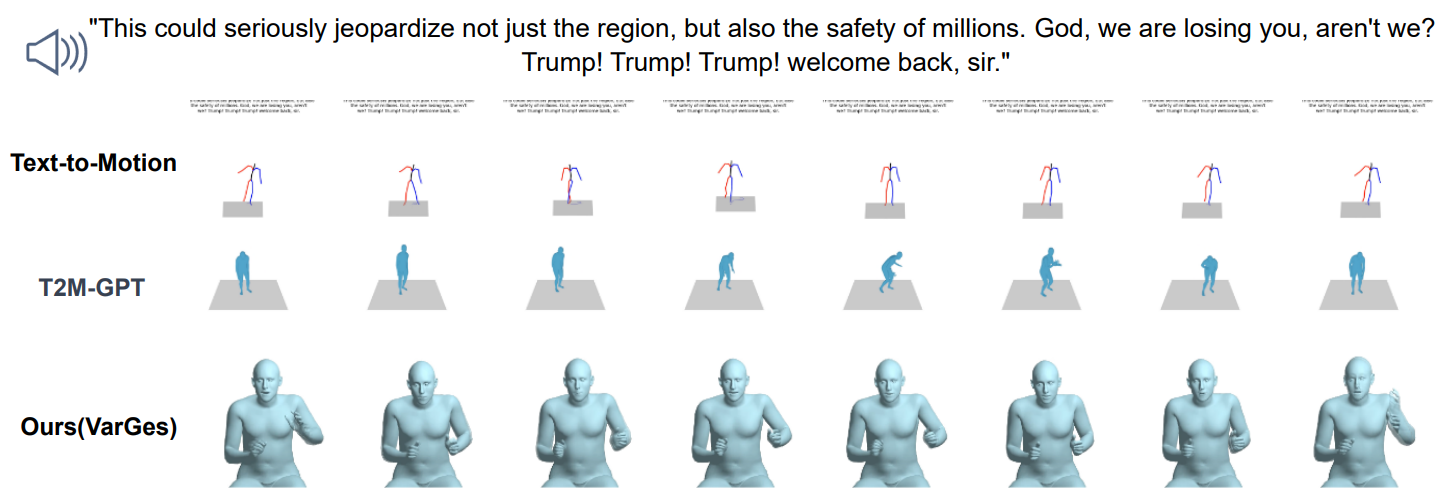}
    \caption{Text-Driven Gesture Generation Comparison. Comparison of gesture generation results using the same text input across different methods.}
    \label{text to gesture}
\end{figure*}

\begin{figure*}[h!t]
    \centering
    \includegraphics[width=1\linewidth]{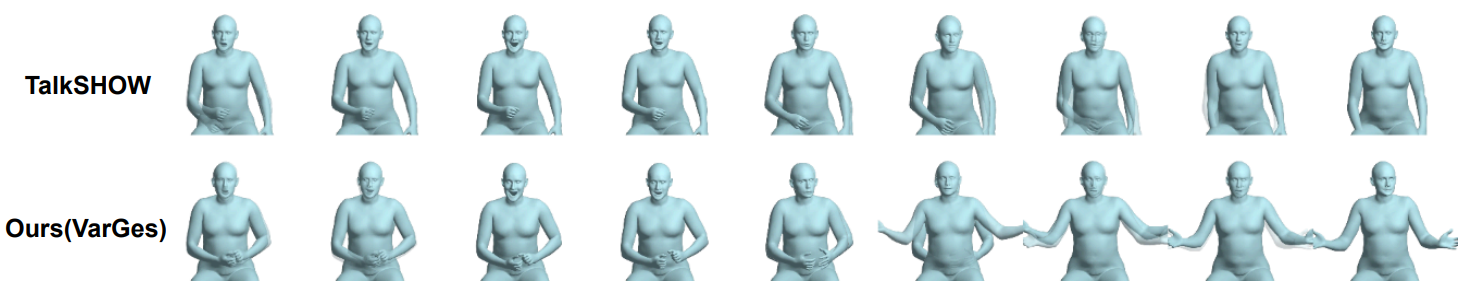}
    \caption{Visualization of 3D gesture variation compared to state-of-the-art method. The co-speech gestures generated by our VarGes method with those produced by the TalkSHOW baseline.}
    \label{Case1}
\end{figure*}

\begin{figure*}[h!t]
    \centering
    \includegraphics[width=1\linewidth]{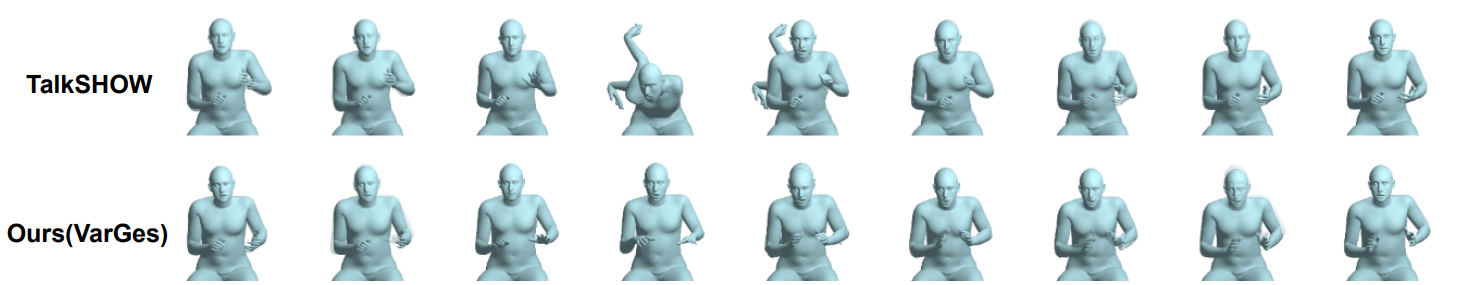}
    \caption{Visualization of gesture naturalness compared to state-of-the-art method. The naturalness and coherence of gestures generated by VarGes against the TalkSHOW method.}
    \label{Case2}
\end{figure*}

\textbf{{Pairwise Preference-Based Comparison.}} {To further validate the effectiveness of VarGes, we conducted a pairwise preference-based evaluation\cite{56}. Specifically, we randomly selected ten sets of video results, each comprising two gesture videos generated by different methods—our method, VarGes, and a baseline method—using the same audio input.}

{Participants were presented with one pair of videos per evaluation page, with the left-right positions of the videos randomized to avoid bias. They were instructed to identify which character’s motions better matched the speech in terms of rhythm, intonation, and meaning. This setup aimed to reduce cognitive load and enable participants to focus on direct comparisons of gesture quality.}

\begin{table}[htbp]
    \centering
    \caption{Preference tests. This table presents the evaluation of our method's performance using pairwise preference-based comparisons.}
    \label{Userstudy_score2}
    \begin{tabular}{@{}ccccccc@{}}
        \hline
        \multirow{2}{*}{Team} & \multicolumn{2}{c}{TalkSHOW} & \multicolumn{2}{c}{\textbf{Ours}} & \multicolumn{2}{c}{Tie} \\ \cline{2-7}
        & Count & Percent & Count & Percent & Count & Percent \\ \hline
        1 & 6 & 0.19 & 24 & 0.77 & 1 & 0.03 \\
        2 & 4 & 0.13 & 26 & 0.84 & 1 & 0.03 \\
        3 & 3 & 0.10 & 27 & 0.87 & 1 & 0.03 \\
        4 & 10 & 0.32 & 20 & 0.65 & 1 & 0.03 \\
        5 & 5 & 0.16 & 23 & 0.74 & 3 & 0.10 \\
        6 & 3 & 0.10 & 26 & 0.84 & 2 & 0.06 \\
        7 & 3 & 0.10 & 24 & 0.77 & 4 & 0.13 \\
        8 & 4 & 0.13 & 25 & 0.81 & 2 & 0.06 \\
        9 & 5 & 0.16 & 25 & 0.81 & 1 & 0.03 \\
        10 & 4 & 0.13 & 25 & 0.81 & 2 & 0.06 \\ \hline
    \end{tabular}
\end{table}

\textbf{{Combined Results and Insights.}} {The results of the MOS-based evaluation (Table~\ref{Userstudy_score}) and pairwise comparisons (Table~\ref{Userstudy_score2}) together provide a robust assessment of VarGes. The MOS ratings highlight the overall superiority of VarGes in human similarity, speech-gesture correlation, smoothness, naturalness, and variation, while the pairwise comparison results (Table~\ref{Userstudy_score2}) demonstrate that VarGes consistently received higher support rates across all ten video sets, with a majority of participants expressing a preference for its generated motions.}

By integrating these two complementary evaluation methods, we ensure a holistic evaluation of VarGes. The MOS protocol quantifies the absolute quality of generated gestures, while the pairwise comparison directly highlights the relative advantages of our approach over baseline methods, offering a nuanced understanding of VarGes's performance.

\subsubsection{Visualization and Case Studies}
We begin by visualizing the 3D meshes generated by our method and comparing them with the original video footage to validate the effectiveness of the StyleCLIPS in qualitative evaluation. As shown in Figure~\ref{Styleclips}, the generated 3D mesh closely mirrors the appearance of the individual in the original video. The key points, including positions and postures, are accurately captured in the 3D mesh, demonstrating strong alignment with the subject's morphology. Moreover, the generated mesh preserves fine details, such as facial expressions and hand gestures, and naturally exhibits different gestures with smooth transitions. These results affirm the accuracy and reliability of our reconstruction process.

{To further validate the effectiveness of our VarGes framework, we introduce two text-driven gesture generation \cite{53,54} comparisons in the visualization section. As shown in Figure~\ref{text to gesture}, the comparison highlights that while text-based approaches can generate diverse gestures, they lack the guidance provided by audio, resulting in silent videos where some actions may appear incongruous. Additionally, the generated gesture sequences often fail to match the duration of the corresponding audio. Furthermore, these methods frequently struggle with fine-grained synchronization, such as aligning lip movements with speech or capturing subtle finger details, as well as maintaining the rhythm of gestures in real-time. In contrast, our method ensures precise lip synchronization and rhythmically coherent gestures, which are essential to achieving natural and expressive human motion generation.}

In addition to the mesh visualization, we compare our method with the previous state-of-the-art (TalkSHOW) using two audio input cases, as illustrated in Figure~\ref{Case1} and Figure~\ref{Case2} respectively. For the first audio input, our method generates gestures with greater variation, particularly in the complexity and richness of hand movements. In the second case, the gestures produced by our method are more natural and coherent, with a significant reduction in issues such as arm jitter and positional misalignment. These comparisons demonstrate that our approach surpasses the baseline in producing natural, fluid, and varied gestures, confirming its effectiveness in generating high-quality co-speech gestures.


\section{Conclusion}
This paper introduces VarGes, a novel framework designed for audio-based 3D human gesture generation with a focus on enhancing gesture variation. At its core, VarGes integrates three synergistic modules: the Variation-Enhanced Feature Extraction (VEFE) module, which integrates {style-reference} videos into a 3D pose estimator to extract StyleCLIPS, {capturing overall motion characteristics such as amplitude, rhythm, and intensity,} thereby enriching input with stylistic nuances; the Variation-Compensation Style Encoder (VCSE), employing a transformer-encoder with an additive attention pooling layer to robustly encode diverse styleclip representations; and the Variation-Driven Gesture Predictor (VDGP), which integrates MFCC audio features and styleclip encodings via cross-attention to modulate a cross-conditional autoregressive model for generating diverse yet natural 3D gestures aligned with audio input. Extensive experimentation on benchmark dataset validates the superiority of our approach in significantly enhancing gesture variation while maintaining naturalness compared to state-of-the-art methods.

\textbf{Limitations and Future Work.} While achieving promising results, our research still faces several limitations. Firstly, the current method is not yet fully optimized for multi-person scenarios and requires further exploration and expansion. Secondly, while we have successfully integrated audio features and stylistic information to enhance the diversity of gesture generation, there is still ample room for further optimizing the balance between diversity and naturalness. Future work will focus on expanding the application to multi-person scenarios, introducing deeper semantic understanding, and further improving the naturalness and diversity of generated gestures.

\section{Acknowledgements}
This work was supported by the Beijing Natural Science Foundation under Grant (4254100), the National Natural Science Foundation of China (11571325, 62441617), the State Key Laboratory of Virtual Reality Technology and Systems, Beihang University (VRLAB2023C04), the Fundamental Research Funds for the Central Universities(CUC2019 A002) and Pubic Computing Cloud, CUC, the Fundamental Research Funds for the Central Universities under Grant (KG16336301), and the China Postdoctoral Science Foundation under Grant (2024M764093).

\section{Declaration of competing interest}
The authors have no competing interests to declare that are
relevant to the content of this article.





\bibliographystyle{IEEEtran}
\bibliography{IEEEabrv,MTT_reveyrand}

\begin{thebibliography}{10}
\providecommand{\url}[1]{#1}
\csname url@rmstyle\endcsname
\providecommand{\newblock}{\relax}
\providecommand{\bibinfo}[2]{#2}
\providecommand\BIBentrySTDinterwordspacing{\spaceskip=0pt\relax}
\providecommand\BIBentryALTinterwordstretchfactor{4}
\providecommand\BIBentryALTinterwordspacing{\spaceskip=\fontdimen2\font plus
\BIBentryALTinterwordstretchfactor\fontdimen3\font minus \fontdimen4\font\relax}
\providecommand\BIBforeignlanguage[2]{{%
\expandafter\ifx\csname l@#1\endcsname\relax
\typeout{** WARNING: IEEEtran.bst: No hyphenation pattern has been}%
\typeout{** loaded for the language `#1'. Using the pattern for}%
\typeout{** the default language instead.}%
\else
\language=\csname l@#1\endcsname
\fi
#2}}

\bibitem{1}
A.~Melinger and W.~J. Levelt, ``Gesture and the communicative intention of the speaker,'' \emph{Gesture}, vol.~4, no.~2, pp. 119--141, 2004.

\bibitem{2}
S.~Goldin-Meadow, ``The role of gesture in communication and thinking,'' \emph{Trends in cognitive sciences}, vol.~3, no.~11, pp. 419--429, 1999.

\bibitem{3}
A.~Kendon, \emph{Gesture: Visible Action as Utterance}.\hskip 1em plus 0.5em minus 0.4em\relax Cambridge: Cambridge University Press, 2004.

\bibitem{4}
Y.~Cheng, P.~Sun, and N.~Chen, ``The essential applications of educational robot: Requirement analysis from the perspectives of experts, researchers and instructors,'' \emph{Computers \& Education}, vol. 126, pp. 399--416, 2018.

\bibitem{5}
J.~Chen, X.~Zhan, Y.~Wang, \emph{et~al.}, ``Medical robots based on artificial intelligence in the medical education,'' in \emph{2021 2nd International Conference on Artificial Intelligence and Education (ICAIE)}.\hskip 1em plus 0.5em minus 0.4em\relax IEEE, 2021, pp. 1--4.

\bibitem{6}
M.~Kipp, ``{Anvil}: The video annotation research tool,'' 2014.

\bibitem{7}
J.~Cassell, H.~H. Vilhj{\'a}lmsson, and T.~Bickmore, ``Beat: the behavior expression animation toolkit,'' in \emph{Proceedings of the 28th annual conference on Computer graphics and interactive techniques}, 2001, pp. 477--486.

\bibitem{18}
J.~Lee and S.~Marsella, ``Nonverbal behavior generator for embodied conversational agents,'' in \emph{International Workshop on Intelligent Virtual Agents}.\hskip 1em plus 0.5em minus 0.4em\relax Springer, 2006, pp. 243--255.

\bibitem{8}
J.~Wagner, T.~Vogt, and E.~Andr{\'e}, ``A systematic comparison of different hmm designs for emotion recognition from acted and spontaneous speech,'' in \emph{Affective Computing and Intelligent Interaction: Second International Conference, ACII 2007 Lisbon, Portugal, September 12-14, 2007 Proceedings 2}.\hskip 1em plus 0.5em minus 0.4em\relax Springer, 2007, pp. 114--125.

\bibitem{9}
S.~Levine, P.~Krähenbühl, S.~Thrun, \emph{et~al.}, ``Gesture controllers,'' in \emph{ACM Siggraph 2010 Papers}, 2010, pp. 1--11.

\bibitem{10}
J.~Li, D.~Kang, W.~Pei, \emph{et~al.}, ``Audio2gestures: Generating diverse gestures from speech audio with conditional variational autoencoders,'' in \emph{Proceedings of the IEEE/CVF International Conference on Computer Vision}, 2021, pp. 11\,293--11\,302.

\bibitem{11}
H.~Yi, H.~Liang, Y.~Liu, \emph{et~al.}, ``Generating holistic 3d human motion from speech,'' in \emph{Proceedings of the IEEE/CVF Conference on Computer Vision and Pattern Recognition}, 2023, pp. 469--480.

\bibitem{12}
L.~Zhu, X.~Liu, X.~Liu, \emph{et~al.}, ``{Taming Diffusion Models for Audio-Driven Co-Speech Gesture Generation},'' in \emph{Proceedings of the IEEE/CVF Conference on Computer Vision and Pattern Recognition}, 2023, pp. 10\,544--10\,553.

\bibitem{13}
I.~Habibie, W.~Xu, D.~Mehta, L.~Liu, H.-P. Seidel, G.~Pons-Moll, M.~Elgharib, and C.~Theobalt, ``Learning speech-driven 3d conversational gestures from video,'' in \emph{Proceedings of the 21st ACM International Conference on Intelligent Virtual Agents}, 2021, pp. 101--108.

\bibitem{14}
T.~Ao, Z.~Zhang, and L.~Liu, ``Gesturediffuclip: Gesture diffusion model with {CLIP} latents,'' in \emph{ACM Trans. Graph.}, 2023.

\bibitem{57}
Z.~Fan, L.~Ji, P.~Xu, F.~Shen, and K.~Chen, ``Everything2motion: Synchronizing diverse inputs via a unified framework for human motion synthesis,'' in \emph{Proceedings of the AAAI Conference on Artificial Intelligence}, vol.~38, no.~2, 2024, pp. 1688--1697.

\bibitem{15}
Y.~Yoon, B.~Cha, J.~Lee, \emph{et~al.}, ``Speech gesture generation from the trimodal context of text, audio, and speaker identity,'' \emph{ACM Transactions on Graphics (TOG)}, vol.~39, no.~6, pp. 1--16, 2020.

\bibitem{16}
U.~Bhattacharya, E.~Childs, N.~Rewkowski, \emph{et~al.}, ``Speech2affectivegestures: Synthesizing co-speech gestures with generative adversarial affective expression learning,'' in \emph{Proceedings of the 29th ACM International Conference on Multimedia}, 2021, pp. 2027--2036.

\bibitem{17}
H.~Liu, Z.~Zhu, N.~Iwamoto, Y.~Peng, Z.~Li, Y.~Zhou, E.~Bozkurt, and B.~Zheng, ``Beat: A large-scale semantic and emotional multi-modal dataset for conversational gestures synthesis,'' in \emph{European conference on computer vision}.\hskip 1em plus 0.5em minus 0.4em\relax Springer, 2022, pp. 612--630.

\bibitem{42}
M.~LOPER, N.~MAHMOOD, J.~ROMERO, \emph{et~al.}, ``{SMPL}: A skinned multi-person linear model,'' \emph{ACM Transactions on Graphics(TOG)}, vol.~34, no.~6, p. 248, 2015.

\bibitem{43}
G.~PAVLAKOS, V.~CHOUTAS, N.~GHORBANI, \emph{et~al.}, ``Expressive body capture: 3d hands, face, and body from a single image,'' in \emph{Proceedings of the IEEE/CVF Conference on Computer Vision and Pattern Recognition}, 2019, pp. 10\,975--10\,985.

\bibitem{44}
H.~Zhang, Y.~Tian, Y.~Zhang, \emph{et~al.}, ``{PYMAF-X}: Towards well-aligned full-body model regression from monocular images,'' \emph{IEEE Transactions on Pattern Analysis and Machine Intelligence}, 2023.

\bibitem{22}
M.~Lhommet, Y.~Xu, and S.~Marsella, ``{Cerebella}: Automatic generation of nonverbal behavior for virtual humans,'' in \emph{Proceedings of the AAAI Conference on Artificial Intelligence}, 2015.

\bibitem{23}
D.~A. Freedman, \emph{Statistical Models: Theory and Practice}.\hskip 1em plus 0.5em minus 0.4em\relax Cambridge: Cambridge University Press, 2009.

\bibitem{24}
M.~Neff, M.~Kipp, I.~Albrecht, \emph{et~al.}, ``Gesture modeling and animation based on a probabilistic re-creation of speaker style,'' \emph{ACM Transactions On Graphics (TOG)}, vol.~27, no.~1, pp. 1--24, 2008.

\bibitem{25}
S.~Levine, C.~Theobalt, and V.~Koltun, ``Real-time prosody-driven synthesis of body language,'' in \emph{ACM SIGGRAPH Asia 2009 Papers}, 2009, pp. 1--10.

\bibitem{26}
S.~Kopp, B.~Krenn, S.~Marsella, A.~N. Marshall, C.~Pelachaud, H.~Pirker, K.~R. Th{\'o}risson, and H.~Vilhj{\'a}lmsson, ``Towards a common framework for multimodal generation: The behavior markup language,'' in \emph{Intelligent Virtual Agents: 6th International Conference, IVA 2006, Marina Del Rey, CA, USA, August 21-23, 2006. Proceedings 6}, 2006, pp. 205--217.

\bibitem{27}
Y.~Ferstl and R.~McDonnell, ``Investigating the use of recurrent motion modelling for speech gesture generation,'' in \emph{Proceedings of the 18th International Conference on Intelligent Virtual Agents}, 2018, pp. 93--98.

\bibitem{28}
T.~Kucherenko, D.~Hasegawa, N.~Kaneko, \emph{et~al.}, ``Moving fast and slow: Analysis of representations and post-processing in speech-driven automatic gesture generation,'' \emph{International Journal of Human–Computer Interaction}, vol.~37, no.~14, pp. 1300--1316, 2021.

\bibitem{29}
K.~Takeuchi, D.~Hasegawa, S.~Shirakawa, \emph{et~al.}, ``Speech-to-gesture generation: A challenge in deep learning approach with bi-directional lstm,'' in \emph{Proceedings of the 5th International Conference on Human Agent Interaction}, 2017, pp. 365--369.

\bibitem{30}
C.~Ahuja, D.~W. Lee, Y.~I. Nakano, and L.-P. Morency, ``Style transfer for co-speech gesture animation: A multi-speaker conditional-mixture approach,'' in \emph{Computer Vision--ECCV 2020: 16th European Conference, Glasgow, UK, August 23--28, 2020, Proceedings, Part XVIII 16}.\hskip 1em plus 0.5em minus 0.4em\relax Springer, 2020, pp. 248--265.

\bibitem{31}
H.~Liu, N.~Iwamoto, Z.~Zhu, \emph{et~al.}, ``{DISCO}: Disentangled implicit content and rhythm learning for diverse co-speech gestures synthesis,'' in \emph{Proceedings of the 30th ACM International Conference on Multimedia}, 2022, pp. 3764--3773.

\bibitem{32}
X.~Liu, Q.~Wu, H.~Zhou, \emph{et~al.}, ``Learning hierarchical cross-modal association for co-speech gesture generation,'' in \emph{Proceedings of the IEEE/CVF Conference on Computer Vision and Pattern Recognition}, 2022, pp. 10\,462--10\,472.

\bibitem{33}
J.~Kim, J.~Kim, and S.~Choi, ``{FLAME}: Free-form language-based motion synthesis \& editing,'' in \emph{Proceedings of the AAAI Conference on Artificial Intelligence}, vol.~37, no.~7, 2023, pp. 8255--8263.

\bibitem{34}
S.~Alexanderson, R.~Nagy, J.~Beskow, \emph{et~al.}, ``Listen, denoise, action! audio-driven motion synthesis with diffusion models,'' \emph{ACM Transactions on Graphics (TOG)}, vol.~42, no.~4, pp. 1--20, 2023.

\bibitem{35}
H.~Xue, S.~Yang, Z.~Zhang, Z.~Wu, M.~Li, Z.~Dai, and H.~Meng, ``Conversational co-speech gesture generation via modeling dialog intention, emotion, and context with diffusion models,'' in \emph{ICASSP 2024-2024 IEEE International Conference on Acoustics, Speech and Signal Processing (ICASSP)}.\hskip 1em plus 0.5em minus 0.4em\relax IEEE, 2024, pp. 8296--8300.

\bibitem{36}
S.~Qian, Z.~Tu, Y.~Zhi, \emph{et~al.}, ``Speech drives templates: Co-speech gesture synthesis with learned templates,'' in \emph{Proceedings of the IEEE/CVF International Conference on Computer Vision}, 2021, pp. 11\,077--11\,086.

\bibitem{37}
T.~Ao, Q.~Gao, Y.~Lou, \emph{et~al.}, ``Rhythmic gesticulator: Rhythm-aware co-speech gesture synthesis with hierarchical neural embeddings,'' \emph{ACM Transactions on Graphics (TOG)}, vol.~41, no.~6, pp. 1--19, 2022.

\bibitem{38}
J.~Xu, W.~Zhang, Y.~Bai, \emph{et~al.}, ``Freeform body motion generation from speech,'' arXiv preprint arXiv:2203.02291, 2022.

\bibitem{39}
P.~J. Yazdian, M.~Chen, and A.~Lim, ``Gesture2vec: Clustering gestures using representation learning methods for co-speech gesture generation,'' in \emph{2022 IEEE/RSJ International Conference on Intelligent Robots and Systems (IROS)}.\hskip 1em plus 0.5em minus 0.4em\relax IEEE, 2022, pp. 3100--3107.

\bibitem{40}
S.~Yang, Z.~Wu, M.~Li, \emph{et~al.}, ``{QPgesture}: Quantization-based and phase-guided motion matching for natural speech-driven gesture generation,'' in \emph{Proceedings of the IEEE/CVF Conference on Computer Vision and Pattern Recognition}, 2023, pp. 2321--2330.

\bibitem{41}
S.~Taylor, J.~Windle, D.~Greenwood, \emph{et~al.}, ``Speech-driven conversational agents using conditional flow-vaes,'' in \emph{Proceedings of the 18th ACM SIGGRAPH European Conference on Visual Media Production}, 2021, pp. 1--9.

\bibitem{45}
A.~Baevski, Y.~Zhou, A.~Mohamed, and M.~Auli, ``wav2vec 2.0: A framework for self-supervised learning of speech representations,'' in \emph{Advances in Neural Information Processing Systems}, vol.~33, 2020, pp. 12\,449--12\,460.

\bibitem{49}
Y.~Deng, J.~Yang, S.~Xu, D.~Chen, Y.~Jia, and X.~Tong, ``Accurate 3d face reconstruction with weakly-supervised learning: From single image to image set,'' in \emph{Proceedings of the IEEE/CVF conference on computer vision and pattern recognition workshops}, 2019, pp. 0--0.

\bibitem{50}
Y.~Feng, V.~Choutas, T.~Bolkart, D.~Tzionas, and M.~J. Black, ``Collaborative regression of expressive bodies using moderation,'' in \emph{2021 International Conference on 3D Vision (3DV)}.\hskip 1em plus 0.5em minus 0.4em\relax IEEE, 2021, pp. 792--804.

\bibitem{51}
H.~Zhang, Y.~Tian, X.~Zhou, W.~Ouyang, Y.~Liu, L.~Wang, and Z.~Sun, ``Pymaf: 3d human pose and shape regression with pyramidal mesh alignment feedback loop,'' in \emph{Proceedings of the IEEE/CVF international conference on computer vision}, 2021, pp. 11\,446--11\,456.

\bibitem{52}
L.-C. Chen, G.~Papandreou, F.~Schroff, and H.~Adam, ``Rethinking atrous convolution for semantic image segmentation,'' \emph{arXiv preprint arXiv:1706.05587}, 2017.

\bibitem{55}
L.~Van~der Maaten and G.~Hinton, ``Visualizing data using t-sne.'' \emph{Journal of machine learning research}, vol.~9, no.~11, 2008.

\bibitem{46}
E.~Ng, H.~Joo, L.~Hu, H.~Li, T.~Darrell, A.~Kanazawa, and S.~Ginosar, ``Learning to listen: Modeling non-deterministic dyadic facial motion,'' in \emph{Computer Vision and Pattern Recognition (CVPR)}, 2022, pp. 20\,395--20\,405.

\bibitem{48}
Z.~Wang, H.~R. Sheikh, A.~C. Bovik, \emph{et~al.}, ``Objective video quality assessment,'' in \emph{The handbook of video databases: design and applications}, 2003, vol.~41, pp. 1041--1078.

\bibitem{56}
T.~Kucherenko*, P.~Wolfert*, Y.~Yoon*, C.~Viegas, T.~Nikolov, M.~Tsakov, and G.~E. Henter, ``Evaluating gesture generation in a large-scale open challenge: The genea challenge 2022,'' \emph{ACM Transactions on Graphics}, vol.~43, no.~3, pp. 1--28, 2024.

\bibitem{53}
C.~Guo, S.~Zou, X.~Zuo, S.~Wang, W.~Ji, X.~Li, and L.~Cheng, ``Generating diverse and natural 3d human motions from text,'' in \emph{Proceedings of the IEEE/CVF Conference on Computer Vision and Pattern Recognition}, 2022, pp. 5152--5161.

\bibitem{54}
J.~Zhang, Y.~Zhang, X.~Cun, Y.~Zhang, H.~Zhao, H.~Lu, X.~Shen, and Y.~Shan, ``Generating human motion from textual descriptions with discrete representations,'' in \emph{Proceedings of the IEEE/CVF conference on computer vision and pattern recognition}, 2023, pp. 14\,730--14\,740.

\end{thebibliography}
%

\end{document}